\documentclass[10pt,twocolumn,letterpaper]{article}

\usepackage{wacv}
\usepackage{times}
\usepackage{epsfig}
\usepackage{graphicx}
\usepackage{amsmath}
\usepackage{amssymb}
\usepackage{booktabs}
\def\wacvPaperID{****} % Enter the WACV Paper ID here

\wacvapplicationstrack % Uncomment this line if you are submitting to the Applications Track.

\wacvfinalcopy % *** Uncomment this line for the final submission

\ifwacvfinal
\usepackage[breaklinks=true,bookmarks=false]{hyperref}
\else
\usepackage[pagebackref=true,breaklinks=true,colorlinks,bookmarks=false]{hyperref}
\fi

\begin{document}

%%%%%%%%% TITLE
\title{Wooden Sleeper Deterioration Detection for Rural Railway Prognostics Using Unsupervised Deeper FCDDs}

\author{Takato Yasuno, Masahiro Okano and Junichiro Fujii\\
Research Institute for Infrastructure Paradigm Shift, Yachiyo Engineering,Co.,Ltd.\\
Taito-ku, Tokyo, 111-8648, Japan\\
{\tt\small \{tk-yasuno,ms-okano,jn-fujii\}@yachiyo-eng.co.jp}
%\and
%2nd\\
%2nd affiliation.\\
%2nd address\\
%{\tt\small ms-okano@yachiyo-eng.co.jp}
}

\maketitle
\thispagestyle{empty}

%%%%%%%%% ABSTRACT
\begin{abstract}
% problem
Maintaining high standards for user safety during daily railway operations is crucial for railway managers. To aid in this endeavor, top- or side-view cameras and GPS positioning systems have facilitated progress toward automating periodic inspections of defective features and assessing the deteriorating status of railway components. However, collecting data on deteriorated status can be time-consuming and requires repeated data acquisition because of the extreme temporal occurrence imbalance. 
% method
In supervised learning, thousands of paired data sets containing defective raw images and annotated labels are required. However, the one-class classification approach offers the advantage of requiring fewer images to optimize parameters for training normal and anomalous features. The deeper fully-convolutional data descriptions (FCDDs) were applicable to several damage data sets of concrete/steel components in structures, and fallen tree, and wooden building collapse in disasters. However, it is not yet known to feasible to railway components.    
% objective
In this study, we devised a prognostic discriminator pipeline to automate one-class damage classification using the deeper FCDDs for defective railway components. We also performed ablation studies of the deeper backbone based on convolutional neural networks (CNNs). Furthermore, we visualized deterioration features by using transposed Gaussian upsampling.
% finding
We demonstrated our application to railway inspection using a video acquisition dataset of railway track from backward view at a cloudy and sunny scene. Finally, we examined the usability of our approach for prognostics and future work on railway inspection.
\end{abstract}

\begin{figure}
\centering
\includegraphics[width=0.45\textwidth]
{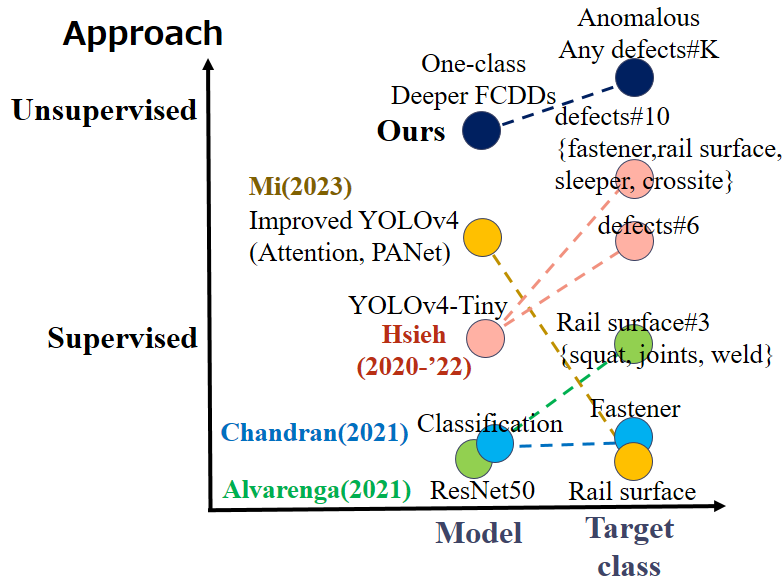}
\caption{Related models and targets for railway prognostic inspection and our approach.}
\label{fig:previousApps}
\end{figure}

%%%%%%%%% BODY TEXT
\section{Introduction}
\subsection{Wooden Sleeper and Derailment Risk}
While analyzing fatal train accidents on Europe's mainline railways from 1980 to 2019, Evans discovered that train collisions, derailments, and railroad crossings were the primary causes of accidents \cite{Evans2011}\cite{Evans2020}. In Japan, according to the traffic safety committee's 1987-2018 derailment statistics, there were 173 cases, with three main causes: 65 natural disasters, 47 railroad crossings, and 33 railway infrastructures \cite{Oyama2022}. To prevent the potential accident of derailment, condition-based and risk-based maintenance in railway infrastructures have been one of key activities for daily operations.
In detail, derailment accidents have been categorized into two types based on the railway tracks: derailment among inter-rails and riding on the outside rail.

Frequently, inter-rail derailments occur because decayed wood sleepers weaken the supporting load of the rail, causing the distance between the parallel rails to expand. Rural railway managers typically operate on a small-scale and have a weak financial status, where inter-rail derailment can negatively impact the profit per day kilometers because of the short length of running operations. Given that rural rail tracks predominantly use wood sleepers, derailment among the inter-rails is more likely to occur. Therefore,  in rural railway maintenance, monitoring wooden sleeper deterioration is critical to reduce the risk of derailment. However, repairing or renewing a decayed sleeper typically requires significant human labor.
To improve the performance of weekly inspections of rural railway tracks, deep learning-based visual inspection techniques can be employed.    

\subsection{Deep Learning for Visual Track Inspection}
Tang et al. reviewed articles on artificial intelligence applications in railway systems from 2010 to 2020, categorizing them into five subdomains: maintenance and inspection, traffic planning and management, safety and security, autonomous driving and control, and passenger mobility \cite{Tang2022}. The most popular research field was found to be maintenance and inspection, with over 81 papers (57\%). 
Ji et al. reviewed existing deep learning applications for rail-track condition monitoring from 2013 to 2021, specifically focusing on supervised deep learning and recent adoption by rail industries \cite{Albert2021}. The number of papers on this topic surged in 2018, and 14 regions worldwide were represented in rail-related research. This indicates that the rail industry has shown growing interest in adopting deep learning methods. Of the studies, 70\% used raw image-type data for deep learning models, while the remaining types included acoustic emission signals, defectograms, maintenance records, and synthetic data from generative models. 
The purpose of these studies was to detect, classify, and/or localize rail surface defects, including various components such as rails, insulators, valves, fasteners, switches, and track intrusions. Over 38 deep learning models have been adopted by researchers, with CNNs being popular for feature extraction and RNN/LSTM being used for sequential data types. Researchers followed a consistent process flow for deep learning applications for rail-track condition monitoring, with the first subsystem being image acquisition (using cameras/recording devices) installed on rail maintenance vehicles to capture input data. The second subsystem involved optional preprocessing, where images were resized, enhanced, noise-removed, or cropped for target areas using image-processing techniques. The input data were then prepared for training and testing deep learning models. Finally, the trained model was produced using the parameters for real-world applications. Given the criticality of rail-track condition monitoring, inspections by human operators could double the accuracy of the system. However, the distribution of rail-track image data is often uneven and extremely disproportional, resulting in class imbalance problems.
This study proposes an unsupervised learning method using a one-class classification algorithm as a novel application in rail-track condition monitoring.
                 
\subsection{One-class Deterioration Detector Application}
Figure \ref{fig:previousApps} depicts several railway inspection applications that utilize deep learning models for the detection of defective classes of railway components. These applications are often based on supervised deep learning approaches such as classification \cite{Alvarenga2021, Chandran2021} and object detection \cite{Mi2023, Hsieh2020, Hsieh2022}. In supervised learning studies, the authors assigned class labels by annotating whole images and bounding boxes that enclosed defective regions. The railway components targeted for defect detection include rail tracks, fasteners, and sleepers.
Railway defects are inherently uncertain events, and the number of anomalous images is often imbalanced toward the normal class. Defect classes are not yet completely defined in railway inspections. For instance, Hsieh et al. \cite{Hsieh2020} defined six normal classes and four defective classes by focusing on clips on wooden/concrete crossties, spikes, fishplate, slide-bed plates, and guard rail plates. The authors collected a limited number of real images, which resulted in seven classes having fewer than 100 defective images and three classes having two or more hundred images. It is not easy to reconstruct synthetic images to represent the health condition of railway components amidst a complex background consisting of trees, grass, and ballast stone. Furthermore, generating defective features as annotation data that can contribute to architectural performance is challenging.
Collecting defective status data to build a railway inspection application always requires significant time investment, given that the temporal occurrence of defects is extremely imbalanced. To achieve stable and high performance, a supervised learning approach demands thousands of paired datasets consisting of defective real images and annotated labels or bounding boxes. 
In contrast, the unsupervised anomaly detection approach has the advantage of requiring fewer images to optimize the parameters for training normal and anomalous features. Moreover, the visual heat map explanation enables us to discriminate between localized defective features. 
The authors \cite{Yasuno2023} found that the deeper fully-convolutional data descriptions (FCDDs) has been applicable to several damage data sets of concrete/steel components in structures: pavement, bridge, and dam, and fallen tree, and wooden building collapse in disasters: typhoon, earthquake. However, it is not yet known to feasible to railway components that includes deterioration of wooden sleepers.
In this study, we propose a prognostic discriminator pipeline to automate the one-class classification of defective railway components. 
%-------------------------------------------------------------------------
\section{Anomaly Detection, Risk-weighted Score}
\subsection{One-class Classification via Deeper FCDDs}
The authors \cite{Yasuno2023} have already formulated the deeper FCDDs and found the applicability to damage data sets of bridge, dam, and building. However, as an unsupervised anomaly detection approach, the deeper FCDDs has been not yet known to feasible to video frame images in front angle of railway track that contains ballast stones, rail, spike, fastener, and concrete/wooden sleepers. 
For risk-based maintenance of rural railways, visualizing hazard-mark heatmaps and computing risk-weighted anomaly scores is crucial for rural railway inspection and prognostic support for effective repairs under usable resources: time, labors, and budget. 

Let $F_i$ be the $i$-th frame of an image with a size of $h\times w$ and let $c$ be the center of the hypersphere boundary between the inlier normal region and the outlier anomalous region. We consider the number of training images and the weight $W$ of the fully convolutional network (FCN). The deep support vector data description (SVDD) objective function \cite{Ruff2018} is formulated as a minimization problem for a deep support vector data description as follows:
\begin{equation}
\min_{W} \frac{1}{n} \sum_{i=1}^{n} \| \Phi^B_W(F_i) - c \|^2,
\end{equation}
where the $\Phi^B_W(F_i)$ denotes a mapping of the deeper CNN to backbone $B$ based on the input frame image. The one-class classification model was formulated using the cross-entropy loss function as follows:
\begin{equation}
\begin{split}
\mathcal{L}_{DeepSVDD} =& - \frac{1}{n} \sum_{i=1}^{n} (1-z_i) \log \ell (\Phi^B_W(F_i)) \\
                    &+ z_i \log [ 1 -  \ell (\Phi^B_W(F_i)) ],
\end{split}
\end{equation}
where $z_i =1$ denotes the anomalous label of the $i$th frame of the image and $z_i =0$ denotes the normal label of the $i$th frame of the image. A pseudo-Huber loss function is introduced to obtain a more robust loss formulation \cite{Ruff2021icml} in Equation (2). Let $\ell(u)$ be the loss function and define the pseudo-Huber loss as follows:   
\begin{equation}
\ell(u) = \exp(-H(u)),~ H(u) = \sqrt{\|u\|^2 + 1} -1.
\end{equation}
Substituting Equation (2) into Equation (3), we obtain the following expression:
\begin{equation}
\begin{split}
(2) \equiv& - \frac{1}{n} \sum_{i=1}^{n} (1-z_i) H (\Phi^B_W(F_i)) \\
                    &+ z_i \log [ 1 -  \exp\{ -H(\Phi^B_W(F_i)) \} ].
\end{split}
\end{equation}
Therefore, a deeper FCDD loss function can be formulated:
\begin{equation}
\begin{split}
&\mathcal{L}_{deeperFCDD} = \frac{1}{n} \sum_{i=1}^{n} \frac{(1-z_i)}{uv} \sum_{x,y} H_{x,y} (\Phi^B_W(F_i)) \\ 
                           &- z_i \log \left[ 1 -  \exp\left\{ \frac{-1}{uv} \sum_{x,y} H_{x,y} (\Phi^B_W(F_i)) \right\} \right],
\end{split}
\end{equation}
where $H_{x,y}(u)$ are the elements $(x,y)$ of the receptive field of size $u\times v$ under a deeper FCDD. 
The risk-weighted anomaly score $S_i^{rw}$ of the $i$th image is expressed as the sum of all the elements of the receptive field as follows:
\begin{equation}
S^{rw}_i(B) = r_i \sum_{x,y} H_{x,y} (\Phi^B_W(F_i)),~i=1,\cdots,n.
\end{equation}
Here, $r_i$ is the weight of the derailment risk caused by the wooden sleeper deterioration. For example, for larger ratios of the curve, the weight can be set higher. 
Specifically, we provided $i$th ratio of the curve to match the GNSS-based position.   
We herein present the construction of a baseline FCDD \cite{Yasuno2023} with an initial backbone $B=0$ and performed CNN27 mapping $\Phi^0_W(F_i)$ from the input frame of image $F_i$ in the dataset. We also present deeper FCDDs focusing on elaborate backbones $B\in \{$VGG16, ResNet101, Inceptionv3$\}$ with a mapping operation $\Phi^{B}_W(F_i)$ to achieve a more robust detection. In this paper, we present ablation studies on a rural railway dataset for detection towards decayed wooden sleeper. 

\subsection{Upsampling Heatmap for Deterioration-mark}
Convolutional neural network (CNN) architectures, comprising millions of common parameters, have exhibited remarkable performance for visual inspection, but the underlying reasons for this superiority remain unclear. Heatmap visualization techniques for detecting and localizing anomalous features are typically categorized as masked sampling and activation map approaches. 
The former includes methods such as occlusion sensitivity \cite{Zeiler2013} and local interpretable model-agnostic explanations \cite{Ribeiro2016}. 
The latter category includes activation maps such as class activation maps (CAMs) \cite{Zhou2015} and gradient-based extensions (Grad-CAM) \cite{Selvaraju2017}. 
Nonetheless, aforementioned methods of disadvantage is its requirement for parallel computation resources and iterative computation time for local partitioning, masked sampling, and for generating a gradient-based heatmap.
In this study for railway inspection applications, we adopt the receptive field upsampling approach \cite{Liznerski2021} to visualize anomalous features using an upsampling-based activation map with Gaussian upsampling from the receptive field of the FCN. The primary advantages of the upsampling approach are the reduced computational resource requirements and shorter computation times. The proposed upsampling algorithm generates a full-resolution anomaly heatmap from the input of a low-resolution receptive field $u\times v$.
 
Let $H\in {R}^{u\times v}$ be a low-resolution receptive field (input), and let $H'\in {R}^{h\times w}$ be a full-resolution of hazard-mark heatmap (output).
We define the 2D Gaussian distribution $G_2(m_1,m_2,\sigma)$ as follows: 
\begin{equation}
\begin{split}
&[G_2(m_1,m_2,\sigma)]_{x,y} \equiv \\
 &\frac{1}{2\pi\sigma^2}\exp\left(-\frac{(x-m_1)^2+(y-m_2)^2}{2\sigma^2}\right).  
\end{split}
\end{equation}
The Gaussian upsampling algorithm from the receptive field is implemented as follows:
\begin{enumerate}
\item $H' \leftarrow 0 \in {R}^{h\times w}$
\item for all output pixels $d$ in $H \leftarrow 0 \in {R}^{u\times v}$
\item \qquad $u(d) \leftarrow$ is upsampled from a receptive field of $d$
\item \qquad $(c_1(u),c_2(u)) \leftarrow$ is the center of the field $u(d)$
\item \qquad $H' \leftarrow H' + d\cdot G_2(c_1,c_2,\sigma)$
\item end for
\item return $H'$ 
\end{enumerate}
After conducting experiments with various datasets, we determined that a receptive field size of $28 \times 28$ is a practical value. When generating a hazardous heatmap, unlike a revealed damage mark, we need to unify the display range that corresponds to the anomaly scores, which range from the minimum to the maximum value. In order to strengthen the defective regions and highlight the hazard marks, we define a display range of [min, max/4], where the quartile parameter is 0.25. This results in the histogram of anomaly scores having a long-tailed shape. If we were to include the complete anomaly score range, the colors would weaken to blue or yellow on the maximum side.
%-------------------------------------------------------------------------
\begin{table}[h]
  \begin{center}
\begin{tabular}{c|c|r|r}
\toprule
Dataset & Size & Normal & Anomalous \\
\midrule
denoised \& cropped & $224^2$ & 3372 & 872 \\
balanced 1~:~1 & $224^2$ & 800 & 872 \\
imbalanced 2~:~1 & $224^2$ & 1600 & 872 \\
imbalanced 3~:~1 & $224^2$ & 2400 & 872 \\
imbalanced 4~:~1 & $224^2$ & 3200 & 872 \\
\bottomrule
\end{tabular}
  \end{center}
\caption{\label{tab:datarail426}Dataset of cloudy railway for ablation studies of class imbalance. Normal images are randomly sampled from 3372.}
\end{table}

\section{Applied Results}

\subsection{Cloudy Track Data Acquisition}
\subsubsection{Data preparedness on cloudy scene}
As presented in Table~\ref{tab:datarail426}, we have demonstrated a railway-related application through an experimental study on a rural railway. We collected the dataset by recording videos using a camera mounted on a train traveling along a single track with a length of approximately 80 km in Japan. The videos were recorded at a rate of 30 frames per second, which provided too much information to be used directly for learning an anomaly detection model. Therefore, we used every fourth frame to generate 27 thousand images, which were then overlapped to represent the railway track in its entirety.

To minimize background noise, we cropped each 4K frame to a size of 1280$\times$2560. We used transfer learning based on ResNet18 and ResNet101 to build two classification models to prepare the input data from the cropped images.
First, since many locations contained large shadows or dark conditions, we built a shadow/dark/without classifier using ResNet18 with three classes: shadows, whole darkness, and without shadows. We randomly selected 3000 images from the cropped images and labeled them as 1458 shadows, 152 whole darkness, and 1390 without shadows. We trained the model using mini-batch 32 and 15 epochs, iterated using Adam, which resulted in a test accuracy of 96.7\%. We predicted the 27 thousand cropped images using the shadow/dark/without classifier and categorized them as 5036 shadow, 1644 whole dark, and 20931 without shadow.

Second, there were many grassy spots on the ballast track, wooden sleepers, and outside the track in each frame image. Therefore, we built a grassy/decayed wooden sleeper/normal classifier using ResNet101 with three classes: grassy, decayed wooden sleeper, and normal without grass. 
We randomly selected 8000 images from the 20931 cropped images without shadows and labeled them as 3372 grassy, 872 decayed wooden sleepers, and 3756 normal without grass. We trained the model using a mini-batch of 128 and 15 epochs, iterated using Adam, which resulted in a test accuracy of 66.4\%. Therefore, we found that even if learning of classification with 3 class, recognizing the deterioration of wooden sleepers was challenging, because of the limited data using 2472 images, and the narrow region of interest per frame, and complex shape and color of wooden deterioration. 
For ablation studies, we accurately labeled 872 decayed wooden sleepers and 800 to 3200 normal wood sleepers without grass. Finally, when we trained our model using both datasets, the input images were resized to $224^2$.

\begin{table}[h]
  \begin{center}
\begin{tabular}{c|c|c|c|c}
\toprule
norm. : anom. & AUC & $F_1$ & Precision & Recall \\
\midrule
1~:~1 & 0.9013 & 0.7412 & 0.8345 & 0.6666 \\
\textbf{2~:~1 } & \textbf{0.9042} & \textbf{0.7520} & \textbf{0.7297} & \textbf{0.7758} \\
3~:~1  &0.9116 & 0.7326 & 0.6850 & 0.7873 \\
4~:~1  &0.9483 & 0.6454 & 0.4939 & 0.9310 \\
\bottomrule
\end{tabular}
  \end{center}
\caption{\label{tab:ablationClass}Imbalance ablation studies on defective detection using our baseline FCDDs for Wooden sleeper (Here, norm. indicates normal, and anom. stands for anomalous.).}
\end{table}

\subsubsection{Ablation studies of class imbalance}
As presented in Table~\ref{tab:ablationClass}, we highlighted class imbalance problem, and implemented ablation studies using our baseline FCDDs with a backbone CNN27 for Wooden sleeper deterioration detection. We provided variation of class imbalance from 1~:~1 to 4~:~1, that means a normal class and an anomalous class by the ratio. In case of 1~:~1, the recall value is relatively worse. Meanwhile, in case of 4~:~1, the precision value is less than 0.5. Further, in case of 3~:~1, the precision value is not better than the case 2~:~1.
In terms of $F_1$, the case 2~:~1 is the highest value, we confirmed that both precision and recall has a better value.  
Thus, we set the ratio of class imbalance 2~:~1, and train the deeper FCDDs for wooden sleeper deterioration detection.

\begin{table}[h]
  \begin{center}
\begin{tabular}{c|c|c|c|c}
\toprule
Backbone & AUC & $F_1$ & Precision & Recall \\
\midrule
CNN27 & 0.9042 & 0.7520 & 0.7297 & 0.7758 \\
VGG16 &0.9660 & 0.8302 & 0.7607 & 0.9137 \\
ResNet101 &0.9681 & 0.8501 & 0.8082 & 0.8965 \\
\textbf{Inceptionv3} & \textbf{0.9664} & \textbf{0.8657} & \textbf{0.8272} & \textbf{0.9080} \\
\bottomrule
\end{tabular}
  \end{center}
\caption{\label{tab:accSleeper426}Backbone ablation studies on defective detection using our proposed deeper FCDDs for imbalanced  2~:~1 dataset.}
\end{table}

\subsubsection{Training anomaly detector and accuracy}
During the training of the anomaly detector, we fixed the input size to $224^2$. To train the model, we set the mini-batch size to 32 and ran 30 epochs. We used the Adam optimizer with a learning rate of 0.0001, a gradient decay factor of 0.9, and a squared gradient decay factor of 0.99. The training images were partitioned at a ratio of 65:15:20 for the training, calibration, and testing images.
As shown in Table~\ref{tab:accSleeper426}, our deeper FCDDs based on VGG16 outperformed the baseline and other backbone-based deeper FCDDs in the rural railway dataset for detection towards decayed wooden sleeper.

\subsubsection{Deterioration-mark heatmaps for prognostics}
\begin{figure}[h]
\centering
\includegraphics[width=0.43\textwidth]{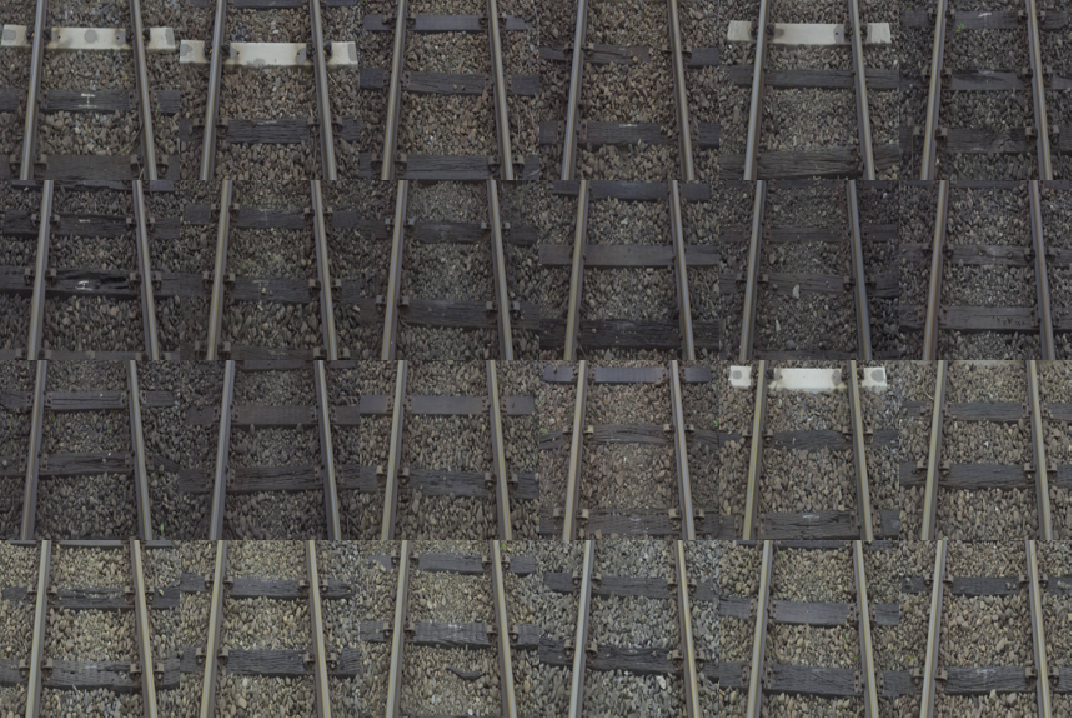} \\
\includegraphics[width=0.43\textwidth]{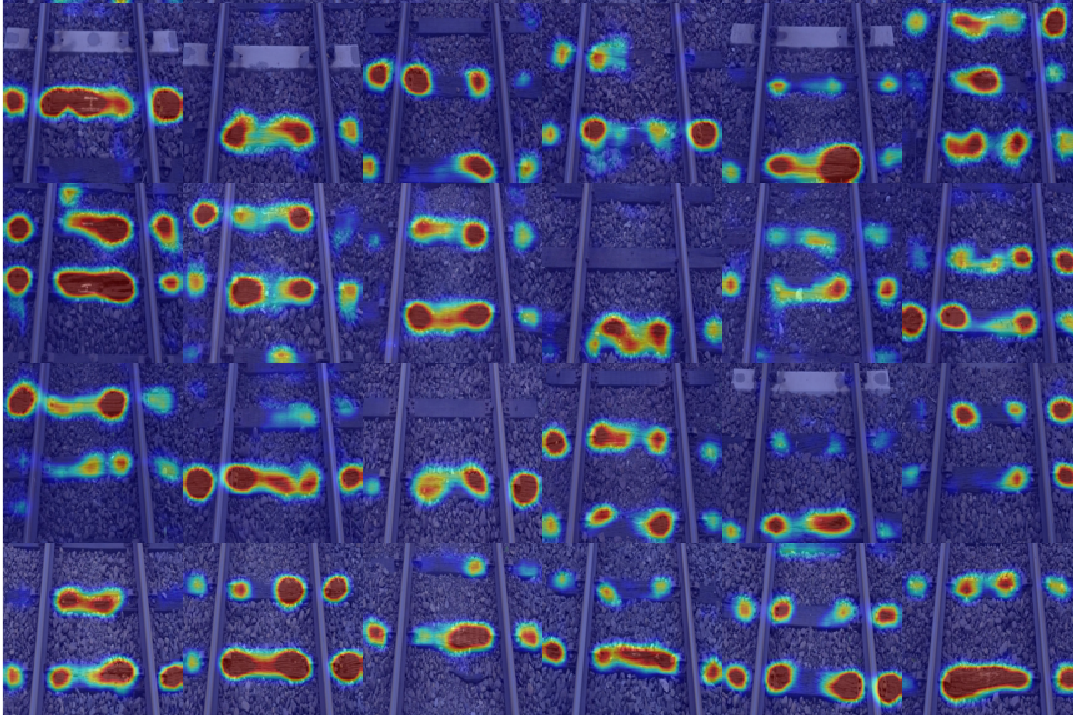}
\caption{\label{fig:rawSleeper426}Input raw images (top) and damage mark heatmaps (bottom) of decayed wooden sleeper using Inceptionv3 backbone.}
\end{figure}

\begin{figure}[h]
\centering
\includegraphics[width=0.40\textwidth]{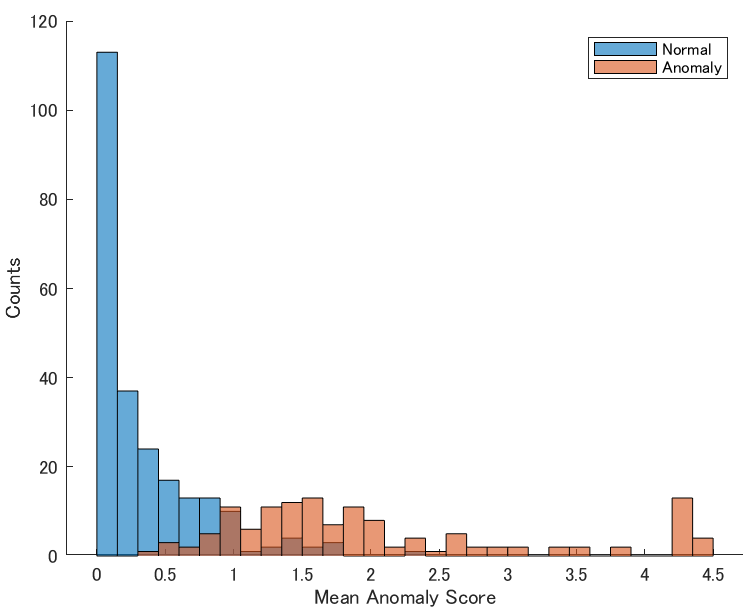}
\caption{\label{fig:histSleeper426}Histogram of decayed wooden sleeper scores corresponding to our deeper FCDD on Inceptionv3 backbone.}
\end{figure}

We visualized he damage features by using Gaussian upsampling of the receptive field in our deeper FCDD network. Additionally, we generated a histogram of the anomaly scores of the test images for the railway-defect dataset. In Figure \ref{fig:rawSleeper426}, a hazard-mark explanation based on a Inceptionv3 backbone is presented. The red region in the heatmap represents the decayed wooden sleepers; however, there are some false negatives because of background noise. Figure \ref{fig:histSleeper426} illustrates that several overlapping bins exist in the horizontal anomaly scores. Therefore, for inspections of decayed wooden sleepers on rural railway tracks, the score range was moderately separated.

% ------------------ results Cloudy 2 : 1
%-------------------------------------------------------------------------
% Sunny Background
\begin{table}[h]
  \begin{center}
\begin{tabular}{c|c|r|r}
\toprule
Dataset & Size & Normal & Anomalous \\
\midrule
denoised \& cropped & $224^2$ & 8097 & 4787 \\
scale 1 & $224^2$ & 2000 & 1000 \\
scale 2 & $224^2$ & 4000 & 2000 \\
scale 3 & $224^2$ & 6000 & 3000 \\
scale 4 & $224^2$ & 8000 & 4000 \\
\bottomrule
\end{tabular}
  \end{center}
\caption{\label{tab:datarail509}Dataset of sunny railway track for ablation studies of data scale. Normal and anomalous images are randomly sampled.}
\end{table}
%-------------------------------------------------------------------------
\subsection{Sunny Track Data Acquisition}
\subsubsection{Data preparedness on sunny scene}
As presented in Table~\ref{tab:datarail509}, we have demonstrated a railway-related application through an experimental study on the rural railway track at another sunny day.  Herein, we used every fourth frame to generate 51 thousand images, which were then overlapped to represent the railway track in its entirety.
To minimize background noise, we cropped each 4K frame to a size of 1280$\times$2560. 
First, we used the aforementioned classifier using ResNet18 with three classes: shadows, whole darkness, and without shadows. 
We have predicted the 51 thousand cropped images using the shadow/dark/without classifier and categorized them as 9970 shadow, 7250 whole dark, and 34249 without shadow.
Second, we also used aforementioned classifier using ResNet101 with three classes: grassy, decayed wooden sleeper, and normal without grass. 
We randomly sampled 16K thousand cropped images from 34249 images without shadow.
And, we have predicted into three classes: 3116 grassy, 4787 decayed wooden sleepers and 8097 normal images without grass. 

\begin{table}[h]
  \begin{center}
\begin{tabular}{c|c|c|c|c}
\toprule
norm. , anom. & AUC & $F_1$ & Precision & Recall \\
\midrule
2K~,~1K & 0.8683 & 0.6986 & 0.6428 & 0.7650 \\
\textbf{4K~,~2K } & \textbf{0.9099} & \textbf{0.7750} & \textbf{0.7104} & \textbf{0.8525} \\
6K~,~3K  &0.9302 & 0.7723 & 0.7465 & 0.8000 \\
8K~,~4K  &0.9182 & 0.7662 & 0.6630 & 0.9075 \\
\bottomrule
\end{tabular}
  \end{center}
\caption{\label{tab:ablationScale}Ablation studies on data scale using our baseline FCDDs for Wooden sleeper (Here, norm. indicates normal, and anom. stands for anomalous.).}
\end{table}

\subsubsection{Ablation studies of data scale}
As presented in Table~\ref{tab:ablationScale}, we highlighted data scale problem, and implemented ablation studies using our baseline FCDDs with a backbone CNN27 for wooden sleeper deterioration detection. 
Based on the aforementioned ablation studies, we set the ratio of class imbalance 2~:~1 in the sunny background dataset.
We provided variation of data scale from 2K~,~1K to 8K~,~4K, that denotes the number of normal and anomalous class. 
In case of 2K~,~1K, any accuracy value is relatively worse. Meanwhile, in case of 8K~,~4K, the precision value is low. Further, in case of 6K~,~3K, the recall value is not better than the case 4K~,~2K.
In terms of $F_1$, the case 4K~,~2K is the highest value, we confirmed that both precision and recall has a better value.  
Thus, we set the data scale 4K~,~2K, and train the deeper FCDDs for wooden sleeper deterioration detection.

\begin{table}[h]
  \begin{center}
\begin{tabular}{c|c|c|c|c}
\toprule
Backbone & AUC & $F_1$ & Precision & Recall \\
\midrule
CNN27 & 0.9099 & 0.7750 & 0.7104 & 0.8525 \\
VGG16 &0.9499 & 0.8213 & 0.7990 & 0.8450 \\
\textbf{ResNet101} & \textbf{0.9525} & \textbf{0.8384} & \textbf{0.8156} & \textbf{0.8625}
\\
Inceptionv3 &0.9464 & 0.8267 & 0.7682 & 0.8950 \\
\bottomrule
\end{tabular}
  \end{center}
\caption{\label{tab:accSleeper509}Backbone ablation studies on defective detection using our proposed deeper FCDDs for data scaled 4000~:~2000 dataset.}
\end{table}

\subsubsection{Training anomaly detector and accuracy}
During the training of the anomaly detector, we fixed the input size to $224^2$. To train the model, we set the mini-batch size to 32 and ran 30 epochs. We used the Adam optimizer with a learning rate of 0.0001, a gradient decay factor of 0.9, and a squared gradient decay factor of 0.99. The training images were partitioned at a ratio of 65:15:20 for the training, calibration, and testing images.
As shown in Table~\ref{tab:accSleeper509}, our deeper FCDDs based on ResNet101 outperformed the baseline and other backbone-based deeper FCDDs in the sunny railway dataset for detection towards decayed wooden sleeper.

\subsubsection{Deterioration-mark heatmaps for prognostics}
\begin{figure}[h]
\centering
\includegraphics[width=0.43\textwidth]{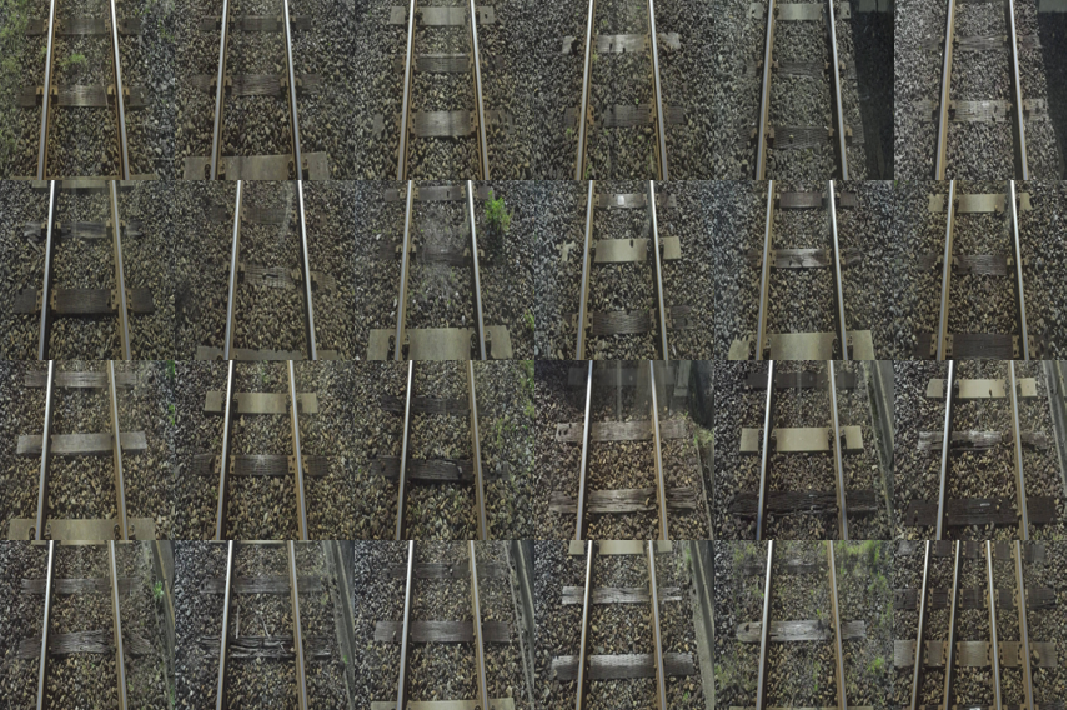} \\
\includegraphics[width=0.43\textwidth]{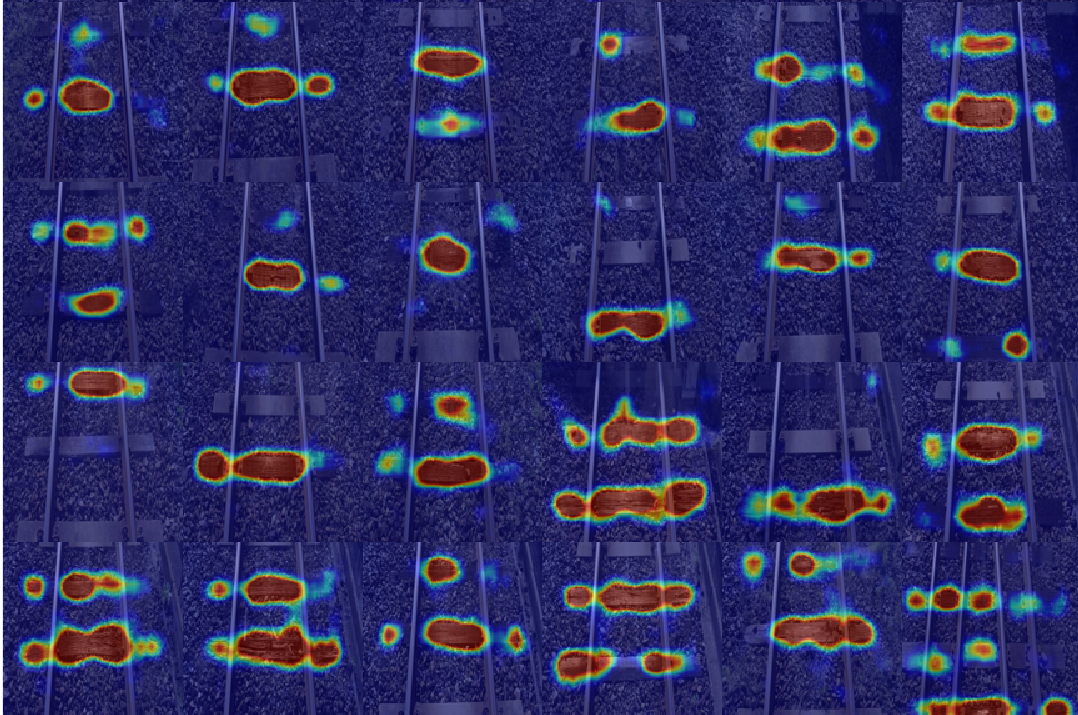}
\caption{\label{fig:rawSleeper509}Input raw images (top) and damage mark heatmaps (bottom) of decayed wooden sleeper using ResNet101 backbone.}
\end{figure}

\begin{figure}[h]
\centering
\includegraphics[width=0.40\textwidth]{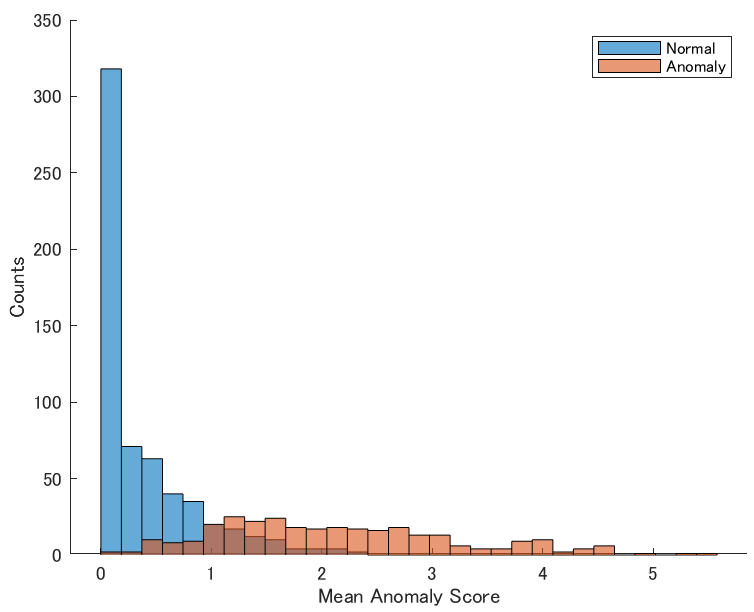}
\caption{\label{fig:histSleeper509}Histogram of decayed wooden sleeper scores corresponding to our deeper FCDD on ResNet101 backbone.}
\end{figure}

We visualized he damage features by using Gaussian upsampling of the receptive field in our deeper FCDD network. Additionally, we generated a histogram of the anomaly scores of the test images for the railway-defect dataset. In Figure \ref{fig:rawSleeper509}, a hazard-mark explanation based on a Inceptionv3 backbone is presented. The red region in the heatmap represents the decayed wooden sleepers; however, there are some false negatives because of background noise. Figure \ref{fig:histSleeper509} illustrates that several overlapping bins exist in the horizontal anomaly scores. Therefore, for inspections of decayed wooden sleepers on rural railway tracks, the score range was moderately separated.

% ------------------ results scale 4K : 2K

%-------------------------------------------------------------------------
%------------------------------------------------------------------------
% Pooled data: cloudy and sunny
\begin{table}[h]
  \begin{center}
\begin{tabular}{c|c|r|r}
\toprule
Dataset & Size & Normal & Anomalous \\
\midrule
cloudy & $224^2$ & 1600 & 872 \\
sunny & $224^2$ & 4000 & 2000 \\
pooled & $224^2$ & 5600 & 2872 \\
\bottomrule
\end{tabular}
  \end{center}
\caption{\label{tab:datarailPool}Pooled dataset of cloudy and sunny scene around railway.}
\end{table}

\subsection{Cloudy-Sunny Pooled Data}
\subsubsection{Data preparedness for pooled dataset}
As presented in Table~\ref{tab:datarailPool}, we have pooled datasets that contains the cloudy and sunny background, where the imbalance ratio is 2~:~1, and better data scale based on this ablation studies.

\begin{table}[h]
  \begin{center}
\begin{tabular}{c|c|c|c|c}
\toprule
Backbone & AUC & $F_1$ & Precision & Recall \\
\midrule
CNN27 & 0.8942 & 0.7523 & 0.7260 & 0.7804 \\
VGG16 &0.9517 & 0.8285 & 0.8003 & 0.8588 \\
ResNet101 &0.9529 & 0.8185 & 0.7453 & 0.9076 \\
\textbf{Inceptionv3} & \textbf{0.9534} & \textbf{0.8295} & \textbf{0.7832} & \textbf{0.8815} \\
\bottomrule
\end{tabular}
  \end{center}
\caption{\label{tab:accSleeperPool}Backbone ablation studies using our proposed deeper FCDDs applied to the cloudy-sunny pooled dataset.}
\end{table}

\subsubsection{Training anomaly detector and accuracy}
As shown in Table~\ref{tab:accSleeperPool}, our deeper FCDDs based on Inceptionv3 outperformed the baseline and other backbone-based deeper FCDDs in the rural railway dataset for detection towards decayed wooden sleeper.

\subsubsection{Deterioration-mark heatmaps for prognostics}
\begin{figure}[h]
\centering
\includegraphics[width=0.43\textwidth]{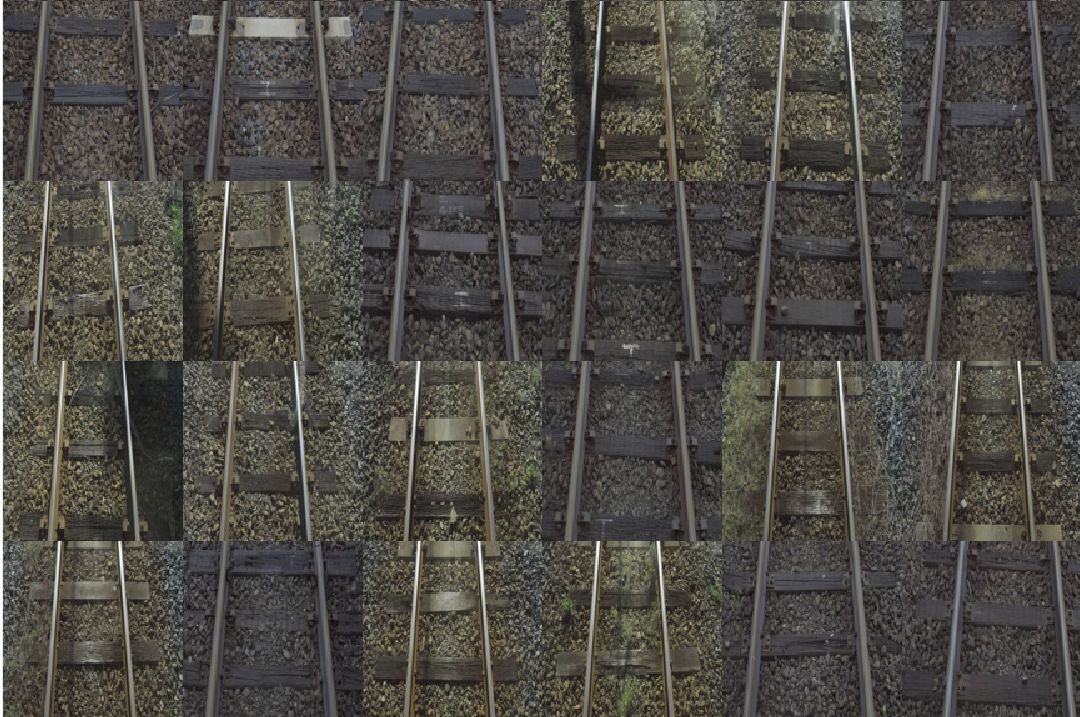} \\
\includegraphics[width=0.43\textwidth]{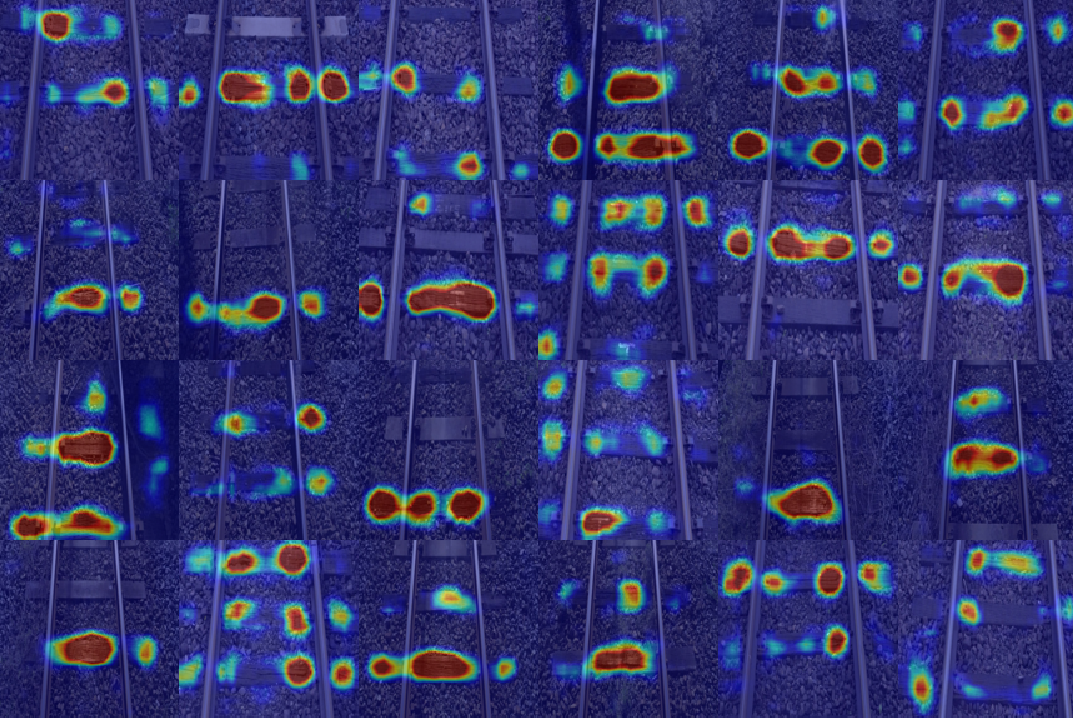}
\caption{\label{fig:rawSleeperPool}Input raw images (top) and damage mark heatmaps (bottom) of decayed wooden sleeper using Inceptionv3 backbone.}
\end{figure}

\begin{figure}[h]
\centering
\includegraphics[width=0.40\textwidth]{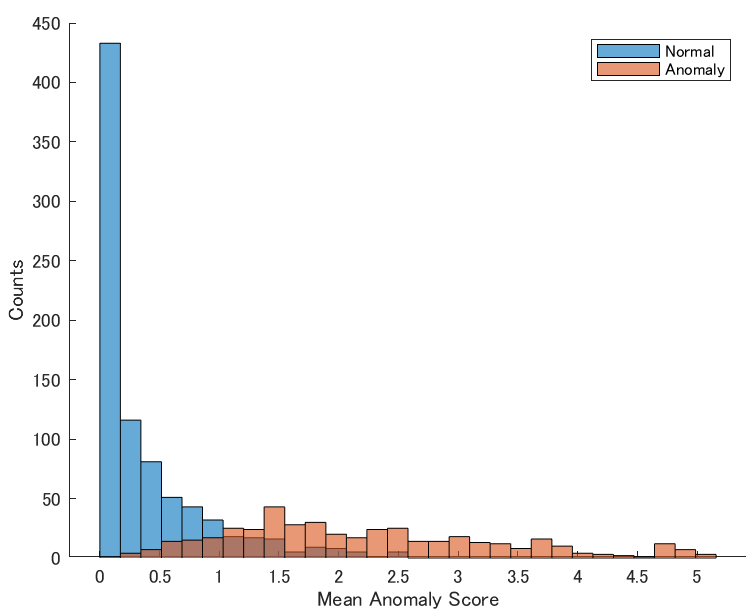}
\caption{\label{fig:histSleeperPool}Histogram of decayed wooden sleeper scores corresponding to our deeper FCDD on Inceptionv3 backbone.}
\end{figure}

We visualized he damage features by using Gaussian upsampling of the receptive field in our deeper FCDD network. Additionally, we generated a histogram of the anomaly scores of the test images for the railway-defect dataset. In Figure \ref{fig:rawSleeperPool}, a hazard-mark explanation based on a Inceptionv3 backbone is presented. The red region in the heatmap represents the decayed wooden sleepers; however, there are some false negatives because of background noise. Figure \ref{fig:histSleeperPool} illustrates that several overlapping bins exist in the horizontal anomaly scores. Therefore, for inspections of decayed wooden sleepers on rural railway tracks, the score range was moderately separated.

% ------------------ results Pooled

% ------------------------ Conclusion
\section{Concluding Remarks}
\subsection{Wooden Sleeper Inspection for Rural Railway}
We developed a railway-purpose application to automate one-class anomaly detection by replicating a baseline FCDD using a light backbone CNN network with 27 layers. To ensure feasibility of a railway application, we assessed an unsupervised approach for deeper FCDDs with pretrained backbones of VGG16, ResNet101, and Inceptionv3, and performed ablation studies by comparing with the baseline FCDD.
In order to support the decision to repair the wooden sleeper and minimize the possibility of derailment for safe railway operations, we proposed an index of the risk-weighted anomaly score. Additionally, we visualized hazard-mark heatmaps using direct Gaussian upsampling of the receptive field of the FCN. We evaluated the deeper FCDDs model on experimental targets, such as decayed wooden sleepers. The heatmap indicates that the hazard marks of the decayed wooden sleeper can cause a spike out of the rail and, in the worst case, a running train could result in derailment.
Our experiments produced a little better accuracies around 79\% for AUC and recall. Deeper FCDDs improved the hazard marks for visual explanation, even without annotating the decayed wooden sleeper regions. We discovered that a hazard-localized approach of deeper FCDDs outperformed the baseline FCDD on rural railway datasets.
Our work presents a new solution for deeper FCDDs that offers a wooden sleeper deterioration detection tool for rural railway inspections with better accuracy and hazard-explainability, providing a novel contribution to the field in rural railway track.

\subsection{Limitations for Target and Season}
% limitations
This study discovered the feasibility of unsupervised deterioration detection highlighting wooden sleeper using the deeper FCDDs in rural railway. However, this scope is too limited for railway inspection to make a decision of repair. We are challenging to detect any deterioration using our application, so another target remains such as fastener, and spike that are pushing out or slipping out. This region of target is smaller and closer, but the impact of defective status is larger to be potential accident of derailment. The key of data preparedness is to divide the cropped image with the size of 1280$\times$2560 into the left-side and right-side of rail with the size of 1280$\times$1280, respectively. 
Though the deterioration scope of wooden sleeper contains both side of rail, but the defective scope of fastener and spike must be recognized as either left-side or right-side of rail.   
In addition, the season of this study was at spring, April and May 2023. Another season, especially winter may influence the background noise: larger shadow, decayed grass, iced, and snowy. For more robustness, we can acquire video at various scene, and the prepared data enable to update the parameters of our anomaly detector.  

\subsection{Future Works for Risk-based Maintenance}
% future works
Several promising directions exist for future research to improve the usability of visual inspection applications. To address the challenges of background noise and imbalanced data, augmentation preprocessing such as mixup, and random erasing can be effective for one-class classification models. However, the imbalance issue remains for infrequent defects such as spikes out of rail from wooden sleepers, cracks of concrete sleepers, and holes on ballast tracks. 
To overcome this challenge, the risk-weighted anomaly score generated by our deeper FCDDs can be used in edge devices for effective data acquisition of rare classes. By collecting only the frames that have hazard marks with significantly higher anomaly scores than a predefined threshold, the data acquisition process can be made more efficient. 
For risk-based maintenance to incorporate the potential hazard of derailment in each track, we are able to add the label of curve ratio \cite{Oyama2022}. Then, we can utilize the risk-weighted anomaly score for supporting to make a decision of priority to repair effectively for sustainable and safety operation in rural railway.     
For prognostic monitoring, we could continue to record the risk-weighted scores every inspection, and we could compare the recent score from the previous scores at each wooden sleeper. For deterioration forecast, we could mine the scores every inspection, and apply hazard models based on the threshold that a railway manager make a priority for repair.     

%%% IMPORTANT - IF accepted, uncomment the Acknowledgment.
%We gratefully acknowledge the conductive comments of the anonymous referees. 
\subsection*{Acknowledgment}
The authors wish to thank MathWorks and Takuji Fukumoto for providing helpful MATLAB resources for Automated Visual Inspection. We also thank Nakasha Creative Co.,Ltd., for providing the opportunity of railway study. 
%-------------------------------------------------------------------------

% -------------------- Reference
{\small
\bibliographystyle{ieee_fullname}
\bibliography{egbib}

\begin{thebibliography}{10}\itemsep=-1pt

\bibitem{Alvarenga2021}
Tiago~A. Alvarenga, Alexandre~L. Carvalho, Leonardo~M. Honorio, Augusto~S.
  Cerqueira, Luciano M.~A. Filho, and Rafael~A. Nobrega.
\newblock Detection and classification system for rail surface defects based on
  eddy current.
\newblock {\em Sensors}, 21(23), 2021.

\bibitem{Chandran2021}
Praneeth Chandran, Johnny Asber, Florian Thiery, Johan Odelius, and Matti
  Rantatalo.
\newblock An investigation of railway fastener detection using image processing
  and augmented deep learning.
\newblock {\em Sustainability}, 13(21), 2021.

\bibitem{Evans2011}
Andrew~W. Evans.
\newblock Fatal train accidents on {E}urope's railways: 1980–2009.
\newblock {\em Accident Analysis and Prevention}, 43(1):391--401, 2011.

\bibitem{Evans2020}
Andrew~W. Evans.
\newblock Fatal train accidents on {E}urope's railways: 1980–2019.
\newblock Technical report, Centre for Transport Studies, Imperial College
  London, 2020.

\bibitem{Hsieh2022}
Chen-Chiung Hsieh, Ti-Yun Hsu, and Wei-Hsin Huang.
\newblock An online rail track fastener classification system based on {YOLO}
  models.
\newblock {\em Sensors}, 22(24), 2022.

\bibitem{Hsieh2020}
Chen-Chiung Hsieh, Ya-Wen Lin, Li-Hung Tsai, Wei-Hsin Huang, Shang-Lin Hsieh,
  and Wei-Hung Hung.
\newblock Offline deep-learning-based defective track fastener detection and
  inspection system.
\newblock {\em Sensors and Materials}, 32(10):3429, 2020.

\bibitem{Albert2021}
Albert Ji, Wai~Lok Woo, Eugene Wai~Leong Wong, and Yang~Thee Quek.
\newblock Rail track condition monitoring: A review on deep learning
  approaches.
\newblock {\em Intelligence and Robotics}, 1(2):151--175, 2021.

\bibitem{Liznerski2021}
Philipp Liznerski, Lukas Ruff, Robert~A. Vandermeulen, Billy~Joe Franks, Marius
  Kloft, and Klaus-Robert Müller.
\newblock Explainable deep one-class classification.
\newblock In {\em The International Conference on Learning
  Representations({ICLR})}, Workshop on Uncertainty and Robustness in Deep
  Learning, 2021.

\bibitem{Mi2023}
Zengzhen Mi, Ren Chen, and Shanshan Zhao.
\newblock Research on steel rail surface defects detection based on improved
  {YOLOv4} network.
\newblock {\em Frontiers in Neurorobotics}, 17, 2023.

\bibitem{MSUC2019}
Masashi Miwa.
\newblock Railway maintenance transformed by {N}umerical {O}ptimizer.
\newblock In {\em Mathematical Systems User Conference 2019}, 2019.

\bibitem{Oyama2022}
Tatsuo Oyama and Masashi Miwa.
\newblock Applying probabilistic mathematical modeling approach and ai
  technique to investigate serious train accidents in japan.
\newblock {\em Sustainability Analytics and Modeling}, 2:2667--2596, 2022.

\bibitem{Ribeiro2016}
Marco~Tulio Ribeiro, Sameer Singh, and Carlos Guestrin.
\newblock "why should i trust you?": Explaining the predictions of any
  classifier.
\newblock In {\em Proceedings of the 22nd {ACM SIGKDD} International Conference
  on Knowledge Discovery and Data Mining}, KDD '16, page 1135–1144.
  Association for Computing Machinery, 2016.

\bibitem{Ruff2018}
Lukas Ruff, Robert Vandermeulen, Nico Goernitz, Lucas Deecke, Shoaib~Ahmed
  Siddiqui, Alexander Binder, Emmanuel M{\"u}ller, and Marius Kloft.
\newblock Deep one-class classification.
\newblock In Jennifer Dy and Andreas Krause, editors, {\em Proceedings of the
  35th International Conference on Machine Learning}, volume~80 of {\em
  Proceedings of Machine Learning Research}, pages 4393--4402. PMLR, 10--15 Jul
  2018.

\bibitem{Ruff2021icml}
Lukas Ruff, Robert~A. Vandermeulen, Billy~Joe Franks, Klaus-Robert Müller, and
  Marius Kloft.
\newblock Rethinking assumptions in deep anomaly detection.
\newblock In {\em The International Conference on Machine Learning ({ICML})},
  Workshop on Uncertainty and Robustness in Deep Learning, 2021.

\bibitem{Selvaraju2017}
Ramprasaath~R. Selvaraju, Michael Cogswell, Abhishek Das, Ramakrishna Vedantam,
  Devi Parikh, and Dhruv Batra.
\newblock {Grad-CAM}: Visual explanations from deep networks via gradient-based
  localization.
\newblock In {\em 2017 {IEEE} International Conference on Computer Vision
  ({ICCV})}, pages 618--626, 2017.

\bibitem{Tang2022}
Ruifan Tang, Lorenzo {De Donato}, Nikola Besinovic, Francesco Flammini,
  Rob~M.P. Goverde, Zhiyuan Lin, Ronghui Liu, Tianli Tang, Valeria Vittorini,
  and Ziyulong Wang.
\newblock A literature review of artificial intelligence applications in
  railway systems.
\newblock {\em Transportation Research Part C: Emerging Technologies},
  140:103679, 2022.

\bibitem{Yasuno2023}
Takato Yasuno, Masahiro Okano, and Junichiro Fujii.
\newblock One-class damage detector using deeper fully convolutional data
  descriptions for civil application.
\newblock {\em Advances in Artificial Intelligence and Machine Learning},
  3(2):996--1011, 2023.

\bibitem{Zeiler2013}
Matthew~D Zeiler and Rob Fergus.
\newblock Visualizing and understanding convolutional networks, 2013.

\bibitem{Zhou2015}
Bolei Zhou, Aditya Khosla, Agata Lapedriza, Aude Oliva, and Antonio Torralba.
\newblock Learning deep features for discriminative localization, 2015.

\end{thebibliography}
}

% ------------------ Appendix
\section{Appendix}
\begin{figure}[h]
\centering
\includegraphics[width=0.25\textwidth]{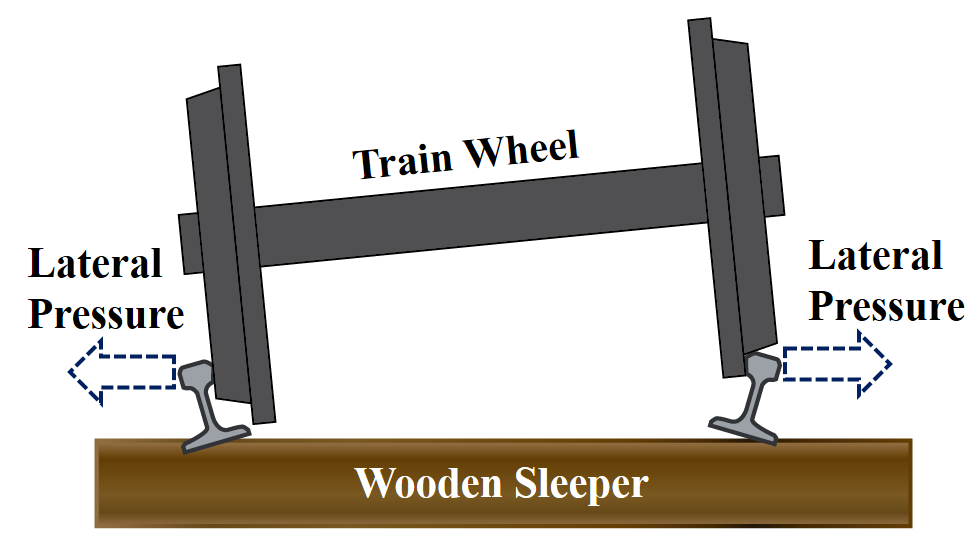}
\caption{\label{fig:derail}Derailment inside track}
\end{figure}

\begin{figure}[h]
\centering
\includegraphics[width=0.25\textwidth]{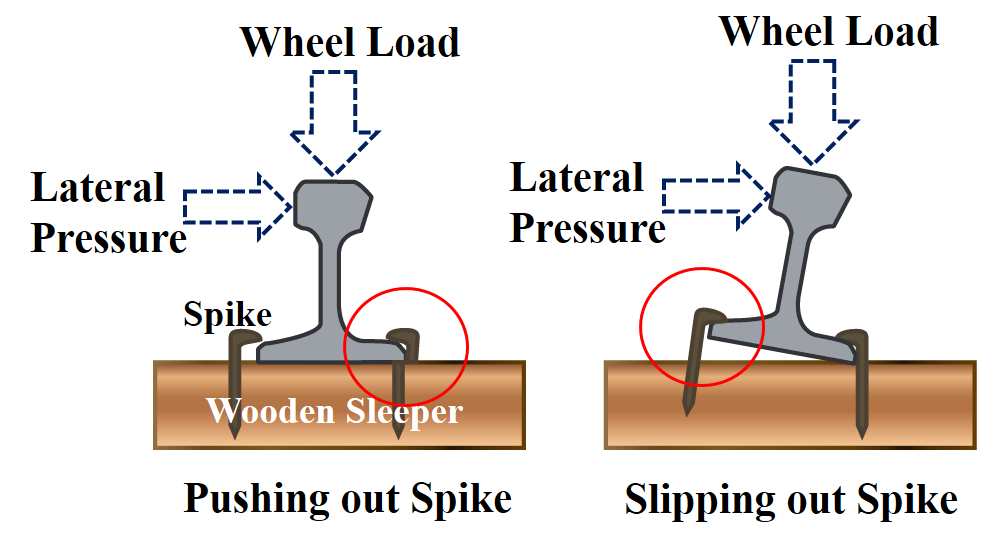}
\caption{\label{fig:spike}Pushing out and slipping out spike on the wooden sleeper}
\end{figure}

Derailment accident has been divided two types : derailment inside track, and riding up outside rail. As shown in Figure~\ref{fig:derail}, that is illustrated with reference to \cite{MSUC2019}, one of cause of derailment inside track is why decayed wooden sleeper makes strengthen the lateral pressure, and the distance between the parallel rails could expand. 
As shown in Figure~\ref{fig:spike}, that is drawn with reference to \cite{MSUC2019}, longitudinal wheel load and  lateral pressure makes pushing out or slipping out the spike on wooden sleeper. Decayed wooden sleeper accelerate to be anomalous status of spike.

\end{document}